\documentclass{uncecomp2025}

\usepackage{graphicx}
\usepackage{caption}
\usepackage{multirow}
\usepackage{amsmath,amssymb,amsfonts,amsthm}
\usepackage{mathrsfs}
\usepackage[title]{appendix}
\usepackage{textcomp}
\usepackage{lmodern}
\usepackage{manyfoot}
\usepackage{booktabs}
\usepackage{algorithm}
\usepackage{algorithmicx}
\usepackage{float}
\floatstyle{plaintop}
\restylefloat{table}
\usepackage{booktabs}
\usepackage{wrapfig}
\usepackage{enumitem}
\usepackage{svg}

\title{Sensitivity Analysis of Image Classification Models using Generalized Polynomial Chaos}

\author{Lukas Bahr$^1$$^,$$^2$, Lucas Poßner$^2$$^,$$^5$, Konstantin Weise$^3$$^,$$^4$, Stephan Matzka$^5$, Sophie Gröger$^6$ and Rüdiger Daub$^7$$^,$$^8$}

\heading{Lukas Bahr, Lucas Poßner, Konstantin Weise, Sophie Gröger and Rüdiger Daub}

\address{
  $^1$
  TUM School of Engineering and Design, Technical University of Munich, Munich, Germany.\\
  e-mail: lukas.bahr@tum.de \and
  $^2$
  Digitization Department for Production, BMW Group, Germany. \and
  $^3$
  Institute for Electrical Power Engineering, Leipzig University of Applied Sciences, Leipzig, Germany. \and
  $^4$
  Methods and Development Group “Brain Networks”, Max Planck Institute for Human Cognitive and Brain Sciences, Leipzig, Germany. \and
  $^5$
  Professorship for Mechatronics and Mechatronic Systems, HTW Berlin, Berlin, Germany. \and
  $^6$
  Professorship of Production Measurement Technology, Technical University of Chemnitz, Chemnitz, Germany.  \and
  $^7$
  Institute for Machine Tools and Industrial Management, Technical University of Munich, Munich, Germany. \and
  $^8$
  Institute for Casting, Composite and Processing Technology (IGCV), Fraunhofer, Augsburg, Germany.
  }


\abstract{
Integrating advanced communication protocols in production has accelerated the adoption of data-driven predictive quality methods, notably machine learning (ML) models.
However, ML models in image classification often face significant uncertainties arising from model, data, and domain shifts. These uncertainties lead to overconfidence in the classification model's output. To better understand these models, sensitivity analysis can help to analyze the relative influence of input parameters on the output.
This work investigates the sensitivity of image classification models used for predictive quality. We propose modeling the distributional domain shifts of inputs with random variables and quantifying their impact on the model's outputs using Sobol indices computed via generalized polynomial chaos (GPC). This approach is validated through a case study involving a welding defect classification problem, utilizing a fine-tuned ResNet18 model and an emblem classification model used in BMW Group production facilities.
}

\begin{document}

\section{INTRODUCTION}
\label{sec:introduction}

Predictive quality has prevailed in the manufacturing industry thanks to the approaches defined in ISO / IEC 31010 \cite{isoiec_risk_2019} and include approaches such as failure mode and effects analysis (FMEA) \cite{iec_iec_2018}, or fault tree analysis \cite{ruijters_fault_2015}. Recent advancements in large language models have enhanced their ability to reason with symbolic data, thereby maintaining the importance of these methods for risk assessment \cite{bahr_knowledge_2024}. 
The rise of the industrial internet of things has enhanced operational efficiency by integrating advanced communication protocols. This development fosters a focus on data-driven predictive quality methods in production processes, such as the application of machine learning (ML) \cite{escobar_quality_2021, mayr_machine_2019}.
Methods such as anomaly detection or classification algorithms can help improve product quality and reliability by finding patterns in data that are not immediately obvious to humans \cite{zonnenshain_quality_2020, possner_anwendung_2024}. In \cite{tercan_machine_2022}, the authors explore common approaches to ML that address predictive quality. Particular image classification models for predicting defects (nOK/OK), such as convolutional neural networks, are dominant in the manufacturing domain.

Understanding the sensitivity of image classification models is of great interest. Sensitivity analysis (SA) studies how the uncertainty of a model's output can be attributed to the uncertainty of its inputs \cite{saltelli_sensitivity_2002}. 
This understanding is crucial because many models, especially those provided by external suppliers, function as black boxes, making it difficult to understand the inner workings. For instance, consider an image classification model that predicts the quality of a weld seam. Image classification models are often poorly calibrated and overconfident on the training data \cite{vasudevan_towards_2019}. Thus, evaluating the sensitivity is essential to determine how changes in input, such as varying lighting conditions in the production environment, affect the model’s reliability.

The challenges for analyzing the sensitivity can be summarized as follows \cite{ovadia_can_2019, gawlikowski_survey_2023}:
\begin{enumerate}[itemsep=0mm]
    \item Many ML models are black-box models, so we cannot explain the predictions given the inputs (model uncertainty).
    \item Errors in the inputs that arise from, for example, inadequate measurements introduce uncertainty in the output (data uncertainty).
    \item Distributional domain shifts in the inputs that arise from, e.g., changes in the production environment, introduce uncertainty in the output (domain uncertainty).
\end{enumerate}

This paper investigates the sensitivity of image classification models to perturbations in the input data. We propose applying geometric transformations to the input images, such as changes in rotation or tilt, or changes on the image channel, such as brightness, treating these variables as uncertain inputs. 
We utilize generalized polynomial chaos (GPC) to approximate a surrogate model of the image classification model's input variables on the output \cite{weise_pygpc_2020}. The surrogate model is then employed to compute the Sobol indices, which quantify the sensitivity with respect to the perturbations and their joint interactions.

In Section~\ref{sec:sa}, we recall the main ideas behind SA. In Section~\ref{sec:method}, we introduce a method to quantify the sensitivity of perturbations on the input image with respect to the output of the classification model. Section~\ref{sec:results} validates the approach through two case studies: (i) a welding seam defect classification problem, utilizing a fine-tuned ResNet18 model and (ii) an actual image classification model for emblem detection employed in BMW Group's production facilities. We conclude with the discussion in Section~\ref{sec:conclusion}.
\section{\MakeUppercase{Sensitivity Analysis}}
\label{sec:sa}

SA aims to quantify the influence of input variables on a model's output \cite{sudret_global_2008, saltelli_sensitivity_2002}. Let $\boldsymbol{y} = f(\boldsymbol{x})$ be a mathematical model with an input vector $\boldsymbol{x} = (x_1, \cdots, x_n)$ of size $n$ and an output vector $\boldsymbol{y} = (y_1, \cdots, y_m)$ of size $m$. We are interested in quantifying the relative importance of how changes in each input $x_i$ influence $y_j$, expressed through relative sensitivity indices \cite{borgonovo_sensitivity_2016}.
Various methods have been developed to analyze a model's sensitivity. The most prominent methods are clustered into local and global SA. 

Local SA methods aim to quantify the influence of individual input features on the model's output by changing specific input values while keeping other inputs constant \cite{saltelli_how_2010}. Local SA approaches are characterized by their low computational cost and ease of application. However, they fail to fully represent nonlinear models and interactions between inputs \cite{tang_comparing_2007}. 
In contrast, global SA methods aim to simultaneously assess the influence of all input variables on the model's output, taking parameter interactions into account \cite{sudret_global_2008}. The oldest global SA methods include regression-based techniques, wherein the regression coefficients are learned from the input variable's influence on the output. The sensitivity can be quantified using measures such as the Pearson correlation coefficient \cite{saltelli_sensitivity_2002}.

However, ML models for image classification are typically nonlinear, allowing the network to learn more complex representations of the input data.
For these nonlinear models, polynomial chaos methods, such as GPC, can be used to approximate a surrogate model to the input-output data, enabling the computation of sensitivity coefficients through easier-to-interpret models \cite{xiu_modeling_2002}. The sensitivity coefficients can then be computed using variance-based methods. Variance-based methods, such as Sobol indices, decompose the variance of the output into the contributions of each input variable. This approach acknowledges that different input variables and their interactions contribute variably to the model's output variations \cite{saltelli_sensitivity_2002-1}.

\subsection{Related work}
\label{subsec:related_research}
Utilizing GPC and Sobol indices for SA has found broad adaptation in various domains such as mechanical, civil, biomedical, and chemical engineering \cite{lucor_sensitivity_2007, ni_using_2019, duong_uncertainty_2016, tosin_tutorial_2020}. In the fields of material engineering and nondestructive testing, GPC has been applied to assess uncertainty in Lorentz force eddy current testing, providing valuable insights into the reliability of defect detection through Sobol indices \cite{weise_uncertainty_2016}. A similar approach has been used to identify optimal stimulation sites for transcranial magnetic stimulation by evaluating sensitivities to uncertain model parameters, such as the conductivity of grey and white matter \cite{saturnino_principled_2019, numssen_determining_2019}.
For analyzing the sensitivity of image models, the authors in \cite{fel_look_2024} propose using Sobol indices to quantify higher-order interactions between input features and the model's output. This approach attributes the neural network's predictions to the most important image regions, outperforming other black-box methods such as RISE (randomized input sampling for explanation) \cite{petsiuk_rise_2018} and even white-box methods like Grad-CAM (gradient-weighted class activation mapping) \cite{selvaraju_grad-cam_2020}.

Building on this literature, the following briefly overviews the theoretical background of GPC and Sobol indices.

\subsection{Generalized polynomial chaos}
\label{subsec:gpc}

GPC is a method used to approximate a surrogate model of $f(\boldsymbol{\xi})$ with random variables $\boldsymbol{\xi}$. The model is approximated by expanding it in terms of multivariate orthogonal polynomial basis functions $\Psi$. This can be expressed as \cite{weise_pygpc_2020, xiu_wieneraskey_2002}
\begin{equation}
\label{eq:pce}
f(\boldsymbol{\xi}) \approx \sum_{\boldsymbol{\alpha} \in A} c_{\boldsymbol{\alpha}} \Psi_{\boldsymbol{\alpha}}(\boldsymbol{\xi}).
\end{equation}
The expansion coefficients $c_{\boldsymbol{\alpha}}$ depend on the multi-index $\boldsymbol{\alpha} = (\alpha_1, \dots, \alpha_d) \in A$ where $A \subseteq \mathbb{N}_0^d$. The multivariate orthogonal polynomial basis functions are given by $\Psi_{\boldsymbol{\alpha}}(\boldsymbol{\xi}) = \prod_{i=1}^d \psi_{\alpha_i}^i(\xi_i)$. These basis functions are products of univariate orthogonal polynomials $\psi_{\alpha_i}(\xi_i)$ of order $\alpha_i$, which are orthogonal with respect to the probability density functions $p_{\Xi_i}(\xi_i)$ \cite{askey_basic_1985}.
In this context, orthogonality is defined by the integral of the inner product of two basis functions, $\psi_{\alpha_i}(\xi_i)$ and $\psi_{\beta_i}(\xi_i)$, weighted by the probability measure $p_{\Xi_i}(\xi_i)$. This integral is zero unless $\alpha_i = \beta_i$. When $\alpha_i = \beta_i$, the integral equals the $L^2$-norm of the basis function. Therefore, the orthogonality condition can be expressed by
\begin{equation}
\label{eq:univariate_poly_orthogonal}
\int \psi_{\alpha_i}(\xi_i) \psi_{\beta_i}(\xi_i) p_{\Xi_i}(\xi_i) \, d\xi_i = \lVert \psi_{\alpha_i}(\xi_i) \rVert^2 \delta_{\alpha_i, \beta_i},
\end{equation}
where $\delta_{\alpha_i \beta_i}$ is the Kronecker delta function \cite{xiu_modeling_2002}. Using the Kronecker delta, which equals one when $\alpha_i = \beta_i$, and applying Equations~\eqref{eq:pce} and \eqref{eq:univariate_poly_orthogonal}, it can be demonstrated that the coefficients $c_{\boldsymbol{\alpha}}$ are expressed as
\begin{equation}
    c_{\boldsymbol{\alpha}}=\frac{1}{\prod_{i} \lVert \psi_{i}(\xi_i) \rVert^2} \int f(\boldsymbol{\xi}) \Psi(\boldsymbol{\xi}) p_\Xi(\boldsymbol{\xi}) \, d\boldsymbol{\xi}.
\end{equation}
The term can be approximated by regression or quadrature methods \cite{xiu_wieneraskey_2002, weise_pygpc_2020}.

By leveraging a surrogate model to approximate a model's output probability distribution using Equation~\eqref{eq:pce}, Sobol indices can be computed efficiently.

\subsection{Sobol indices}
\label{subsec:sobol}

Sobol indices are a variance-based sensitivity analysis method that quantifies the influence of a subset $\boldsymbol{\xi}_\tau$ of a model's inputs on its outputs. The indices $\tau \in \mathcal{P}(\{1, \dots, d\})$ include all interactions of a model's input variables. Thus, Sobol indices can quantify the influence of individual input variables and the combined effect of different interactions of input variables \cite{sobol_global_2001}. We distinguish between total Sobol indices $D_\tau$ and relative Sobol indices $S_\tau$. The total Sobol indices are given by
\begin{equation}
    \label{eq:total_sobol_indices}
    D_\tau=\mathbb{V}_{\boldsymbol{\xi}_\tau}[f(\boldsymbol{\xi})],
\end{equation}
where $\mathbb{V}[f(\boldsymbol{\xi})]$ is the variance of $f(\boldsymbol{\xi})$ explained by the input variables $\boldsymbol{\xi}_\tau$.
For the relative Sobol indices, we normalize Equation~\eqref{eq:total_sobol_indices} by the sum over all interactions of input variables \cite{sudret_global_2008}
\begin{equation}
    \label{eq:relative_sobol_indices}
    S_\tau = \frac{D_\tau}{\sum_\tau D_\tau} = \frac{\mathbb{V}_{\boldsymbol{\xi}_\tau}[f(\boldsymbol{\xi})]}{\sum_\tau \mathbb{V}_{\boldsymbol{\xi}_\tau}[f(\boldsymbol{\xi})]} = \frac{\mathbb{V}_{\boldsymbol{\xi}_\tau}[f(\boldsymbol{\xi})]}{\mathbb{V}[f(\boldsymbol{\xi})]}.
\end{equation}
Thus, $S_\tau$ can be seen as a measure of the relative influence of $\boldsymbol{\xi}_\tau$ on $f(\boldsymbol{\xi})$. Global and relative Sobol indices can be computed using the coefficients in Equation~\eqref{eq:pce} by
\begin{equation}
\label{eq:sobol_indices}
S_\tau = \frac{\sum_{\boldsymbol{\alpha} \in A_\tau} c_{\boldsymbol{\alpha}}^2 \lVert \Psi_{\boldsymbol{\alpha}} \rVert^2}{\sum_{\boldsymbol{\alpha} \in A} c_{\boldsymbol{\alpha}}^2 \lVert \Psi_{\boldsymbol{\alpha}} \rVert^2},
\end{equation}
with $A_\tau = \{\boldsymbol{\alpha} \in A : \alpha_i > 0 \iff i \in \tau\}$ representing the set of multi-indices $\boldsymbol{\alpha}$ for which $\alpha_i > 0$ if and only if $i$ is an element of the subset $\tau$ \cite{sudret_global_2008}.

\section{\MakeUppercase{GPC-SA for Image Classification Models}}
\label{sec:method}  

\begin{figure}[t]
\centering
\includegraphics[width=0.8\textwidth]{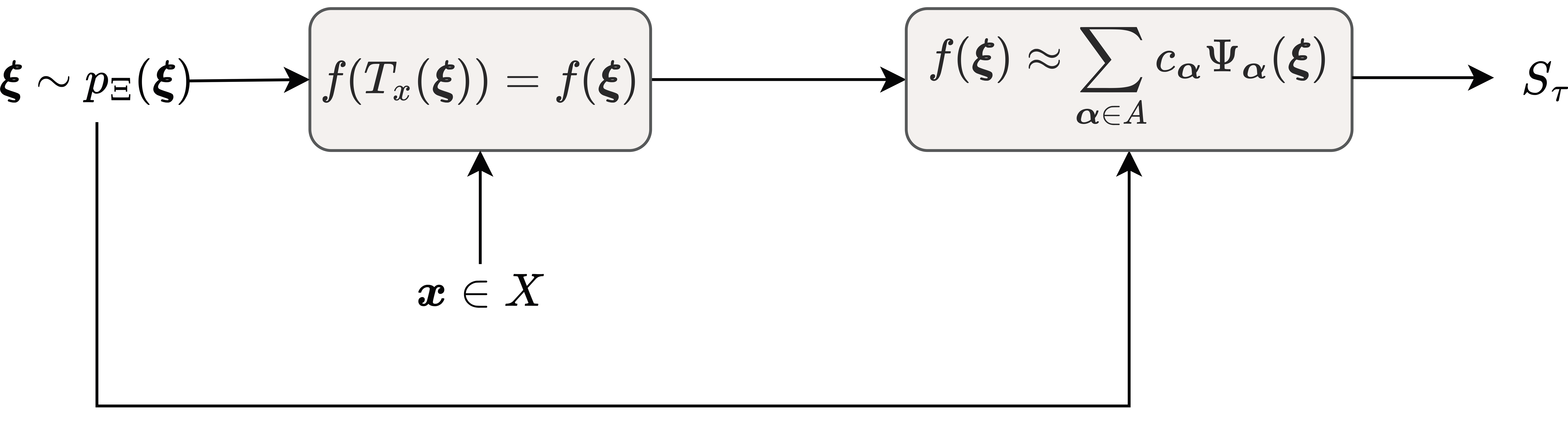}
\caption[]{The fixed input $\boldsymbol{x}$, given the perturbation $\boldsymbol{\xi}$, is transformed using the transformation function $T(\boldsymbol{x}, \boldsymbol{\xi})$ and passed through the black-box model $f$ to obtain the probability distribution $p_Y(\boldsymbol{y} \mid \boldsymbol{\xi})$. 
GPC approximates a surrogate model of $f(T(\boldsymbol{x}, \boldsymbol{\xi}))$ in terms of a series expansion of multivariate orthogonal polynomials.
The surrogate model is then used to compute the Sobol indices $S_\tau$.}
\label{fig:workflow}
\end{figure}

Perturbations on images, such as noise, lighting variations (e.g., brightness), and geometric distortions (e.g., rotation), can significantly impact the performance of image classification models. These perturbations introduce variations in the input data distribution that the model may not be adequately trained on. As a result, the model must account for these variations to maintain robust performance.

Let $f: \mathbb{R}^{m \times n \times c} \rightarrow \{1, \dots, K\}$ define the image classification model with input images $\boldsymbol{x} \in \mathbb{R}^{m \times n \times c}$, where $m$ and $n$ depict the width and height of the image, and $c$ represents the number of image channels. The set $y = \{1, \dots, K\}$ represents the output labels with an associated probability distribution $p_Y(\boldsymbol{y})$ for $K$ classes. 
Now, let $p_Y(\boldsymbol{y} \mid \boldsymbol{x}) = f(\boldsymbol{\theta}, \boldsymbol{x})$ be a probabilistic classification model with learnable parameters $\boldsymbol{\theta}$ that maps $\boldsymbol{x}$ to $p_Y(\boldsymbol{y} \mid \boldsymbol{x})$ \cite{goodfellow_deep_2016}. We aim to quantify the influence of perturbations on the image as transformations of the model's input variables $\boldsymbol{x}$ on its output $p_Y(\boldsymbol{y} \mid \boldsymbol{x})$ using Sobol indices of different interaction orders.

To quantify the relative influence of perturbations on the model's output, we consider a set of random variables $\boldsymbol{\xi} = (\xi_1, \dots, \xi_d)$ drawn from a probability distribution $p_\Xi(\boldsymbol{\xi})$. We define the transformation function $T(\boldsymbol{x}, \boldsymbol{\xi})$ to analyze these perturbations.
We then compute the transformed inputs $\boldsymbol{x'}^{(i)} = T(\boldsymbol{x}, \boldsymbol{\xi}^{(i)})$ for each sample $\boldsymbol{\xi}^{(i)}$. The model then predicts the probability $f(\boldsymbol{x'}^{(i)}) = f(T(\boldsymbol{x}, \boldsymbol{\xi}^{(i)})) = p_Y(\boldsymbol{y} \mid \boldsymbol{x}'^{(i)})$ of $\boldsymbol{x'}$ belonging to class $\boldsymbol{y}$.
By fixing $\boldsymbol{x}$, we define $T_x(\boldsymbol{\xi})$ as a function that depends only on the transformations $\boldsymbol{\xi}$ and serves as an uncertain input of the model. Thus, for a fixed $\boldsymbol{x} \in X$, it follows
\begin{equation}
f(\boldsymbol{x'}) = f(T_x(\boldsymbol{\xi})) = f(\boldsymbol{\xi}) = p_Y(\boldsymbol{y} \mid \boldsymbol{\xi}).
\label{eq:classification_trans_func}
\end{equation}

This approach allows us to isolate and analyze the influence of the perturbations $\boldsymbol{\xi}$ on the model's predictions, facilitated by the computation of Sobol indices to quantify the influence of $\boldsymbol{\xi}$ on the model's output. However, the range of polynomials is $(-\infty, +\infty)$, while the range of a discrete probability distribution is $[0, 1]$. 
To account for this, we apply the logit function in order to map the probability to $(-\infty, +\infty)$. This can be written as
\begin{equation}
f(\boldsymbol{\xi}) = \text{logit}(p_Y(\boldsymbol{y} \mid \boldsymbol{\xi})) = \log\left(\frac{p_Y(\boldsymbol{y} \mid \boldsymbol{\xi})}{1 - p_Y(\boldsymbol{y} \mid \boldsymbol{\xi})}\right).
\label{eq:logit}
\end{equation}

Using Equation~\eqref{eq:pce}, we approximate the surrogate model by
\begin{equation}
f(\boldsymbol{\xi}) = \text{logit}(p_Y(\boldsymbol{y} \mid \boldsymbol{\xi})) \approx \sum_{\boldsymbol{\alpha} \in A} c_{\boldsymbol{\alpha}} \Psi_{\boldsymbol{\alpha}}(\boldsymbol{\xi}).
\end{equation}
Next, we compute the Sobol indices $S_\tau$ for various interaction orders $\tau$ for each input $\boldsymbol{x}'^{(i)}$ using Equation~\eqref{eq:sobol_indices}. To interpret the probabilities, we apply the inverse of Equation~\eqref{eq:logit}, which is the logistic function. This can be written as
\begin{equation}
\text{logistic}\left( \sum_{\boldsymbol{\alpha} \in A} c_{\boldsymbol{\alpha}} \Psi_{\boldsymbol{\alpha}}\right) = \frac{1}{1 + \exp\left(- \sum_{\boldsymbol{\alpha} \in A} c_{\boldsymbol{\alpha}} \Psi_{\boldsymbol{\alpha}}\right)}.
\end{equation}
A schema of the complete approach is provided in Figure~\ref{fig:workflow}.
\section{RESULTS}
\label{sec:results}
In the following, we validate our proposed approach using image classification models that have been fine-tuned on two different datasets. We demonstrate that approximating $f(T(\boldsymbol{x}, \boldsymbol{\xi}))$ using GPC and computing the Sobol indices can be used to quantify the sensitivity of perturbations in the input image on the model output. In Section~\ref{subsec:welding}, we analyze the sensitivity of a classification model trained on a benchmark welding seam dataset. Section~\ref{subsec:emblem} analyzes an image classification model used in BMW Group's vehicle manufacturing plants to detect the correct assembly of brand emblems.

\subsection{TIG welding seam classification}
\label{subsec:welding}

\begin{wraptable}{r}{0.5\textwidth}
\centering
\begin{tabular}{@{}l|lll@{}}
\toprule
           & brightness & rotation   & tilt           \\ \midrule
brightness & 0.171      & 0.03       & 0.162         \\
rotation   & -          & 0.11       & 0.096  \\
tilt       & -          & -          & \textbf{0.37}          \\ \bottomrule
\end{tabular}
\caption[]{Sobol indices from the GPC surrogate model for \textit{brightness}, \textit{rotation}, and \textit{tilt}. The results for all three transformations have a relative influence of 
$0.061$. \textit{Tilt} exhibits the highest influence.}
\label{tab:sobol_welding}
\end{wraptable}

We validate our approach on an image classification model using a pre-trained and fine-tuned ResNet18 model \cite{he_deep_2016}. The model comprises 11.7 million parameters, ReLU activations, and a single 7x7 convolution layer with pooling. The pre-trained ResNet18 model is fine-tuned on benchmark Aluminum 5083 TIG welding seam images recorded with an HDR camera, down-sampled to 224x224 pixels with a single channel and without applying any transformations \cite{bacioiu_automated_2019}.
The training dataset contains 26,666 images, and the test dataset contains 6,588 images. The images show six defect types: good weld, burn-through, contamination, lack of fusion, misalignment, and lack of penetration.

\begin{figure}[t]
\centering
\includegraphics[width=1\textwidth]{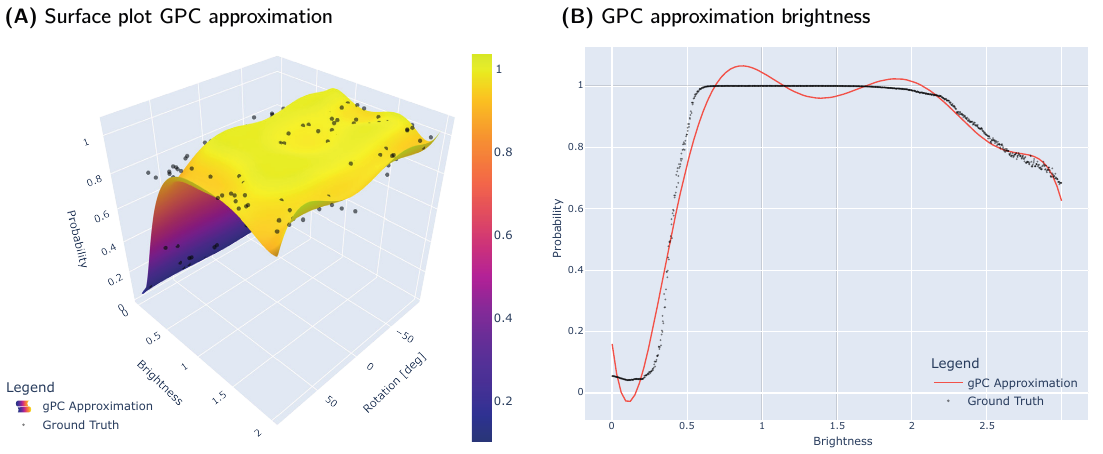}
\caption[]{Figure (A) presents a surface plot of the GPC approximation, illustrating the relationship between \textit{brightness} and \textit{rotation} for the prediction probability, with \textit{tilt} held constant at $0$. Figure (B) displays the approximation of \textit{brightness} while both \textit{rotation} and \textit{tilt} are fixed at $0$.}
\label{fig:gpc_approximation}
\end{figure}

We are interested in evaluating the sensitivity of the fine-tuned image classification model to variations in brightness, rotation, and tilt, as well as the interactions among these transformations. To achieve this, we compute the Sobol indices, which quantify the contribution of each input variable and their interactions with the output variance.
By analyzing the Sobol indices, we can determine the sensitivity of the classification model to each transformation and understand how these transformations interact to affect the model's performance. 

For this purpose, we draw $n=1000$ samples using Latin hypercube sampling (LHS) from a Beta distribution with specified limits for each transformation parameter. The limits are $[0, 2]$ for \textit{brightness} (representing the scaling factor of brightness change, where $1$ represents the original brightness), $[-30, 30]$ for \textit{rotation} (the angle of rotation in degrees), and $[-20, 20]$ for \textit{tilt} (the angle of tilt in degrees).

We apply these transformations to a fixed input image and compute the output probabilities of the image classification model using Equation~\eqref{eq:classification_trans_func}.
The surrogate model is approximated using GPC with Jacobi polynomials. The relative error of the approximation is $0.057$.
We set the maximum total order of polynomials to four and use the Moore-Penrose pseudo inverse to compute the polynomial coefficients. We then determined the Sobol indices with the computed coefficients, as given by Equation~\eqref{eq:sobol_indices}.

Table~\ref{tab:sobol_welding} presents the resulting Sobol indices. These indices indicate that \textit{tilt} has the highest influence on the model, with a value of $0.37$. In contrast, \textit{brightness} and \textit{rotation} exert the lowest influence on the model, each with a value of $0.03$. The combined relative influence of \textit{brightness}, \textit{rotation}, and \textit{tilt} is $0.061$\footnote{For those interested in reproducing the results or exploring the SA approach further, the code including the fine-tuned ResNet18 model is available on GitHub at \url{https://github.com/lpossner/GPC_Image_Classification}.}.

The surrogate models illustrating the GPC approximation for \textit{rotation} and \textit{brightness} are depicted in Figure~\ref{fig:gpc_approximation}. Figure~\ref{fig:gpc_approximation} (A) presents a surface plot of the surrogate model with respect to \textit{rotation} and \textit{brightness}, where \textit{tilt} is fixed at a value of $0$. Figure~\ref{fig:gpc_approximation} (B) shows a plot of the surrogate model with respect to \textit{brightness}, where \textit{rotation} and \textit{tilt} are fixed at a value of $0$.

\subsection{Vehicle emblem classification}
\label{subsec:emblem}

\begin{figure}[t]
\centering
\includegraphics[width=\textwidth]{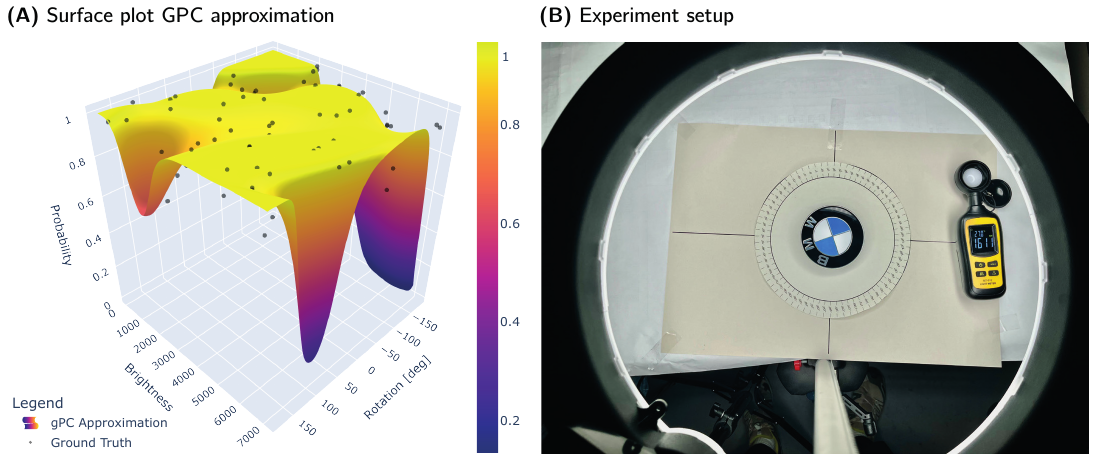}
\caption[]{Figure (A) presents a surface plot of the GPC approximation, illustrating the relationship between \textit{brightness} and \textit{rotation} for the prediction probability. Figure (B) shows the experimental setup used to obtain the validation images.}
\label{fig:emblem_experiment}
\end{figure}

In this experiment, we validate our SA approach on an image classification model fine-tuned to detect the correct assembly of brand emblems in BMW's Group vehicle production plants. The model classifies images into three distinct emblem types and includes an additional class for unclassified instances. Operating as a black box, the image classification model provides output class probabilities for given inputs. We aim to evaluate its sensitivity to variations in \textit{rotation} and \textit{brightness} and the interactions between these transformations, using Sobol indices.

\begin{wraptable}{r}{0.5\textwidth}
\centering
\begin{tabular}{@{}l|lll@{}}
\toprule
           & brightness & rotation \\ \midrule
brightness & 0.229      & \textbf{0.403} \\
rotation   & -          & 0.368 \\ \bottomrule
\end{tabular}
\caption[]{Result of the Sobol indices for \textit{brightness} and \textit{rotation}. The joint interaction exhibits the highest influence on the emblem classification model.}
\label{tab:sobol_emblem}
\end{wraptable}

To conduct our experiment, we set up an apparatus comprising a Keyence camera, an adjustable light source, a luminance meter, and an angle scale. We then adjusted the \textit{rotation} and \textit{brightness} for $n=100$ samples using LHS drawn from a Beta distribution. The limits for the \textit{rotation} transformation (the angle of rotation in degrees) are set between $[-180, 180]$, and for the \textit{brightness} transformation (representing brightness in Lux), the range is $[0, 7000]$. With this configuration, we adjust the \textit{rotation} angle and \textit{brightness} and shot the images (cf. Figure~\ref{fig:emblem_experiment} (B)) to compute the output class probabilities using Equation~\eqref{eq:classification_trans_func}. To approximate the surrogate model, we employ GPC with Jacobian polynomials and subsequently compute the Sobol indices. The relative error obtained from this method is $0.23$, given a maximum total polynomial order of $[6,5]$.

Table~\ref{tab:sobol_emblem} presents the Sobol indices obtained from our experiment. The results indicate that the joint interaction of \textit{brightness} and \textit{rotation} has the highest influence, with a Sobol index of $0.403$. Figure~\ref{fig:emblem_experiment} (A) displays a surface plot of the GPC approximation. In the following chapter, we contextualize the results.

\section{DISCUSSION AND CONCLUSION}
\label{sec:conclusion}

This paper presents preliminary results on the effectiveness of sensitivity analysis of image classification models with perturbed images using GPC. We introduce a method where a fixed input image is transformed by samples from a known distribution that is associated with a family of orthogonal polynomial functions. Using the output of the image classification model for the transformed input, we employ GPC to approximate a surrogate model \cite{weise_pygpc_2020}. This surrogate model enables us to determine the relative influence of each transformation applied to the input on the model's output, quantified as Sobol indices.

In Section~\ref{sec:results}, we demonstrate the approach on an Aluminum 5083 TIG welding seam images dataset for which we fine-tuned a ResNet18 model and on an emblem image classification model utilized in BMW Group's production facilities \cite{bacioiu_automated_2019}. The experiment provides initial results on evaluating the models' sensitivity to images that are exposed to perturbations, which can potentially lower the model's prediction performance. 
Our findings show we can approximate the nonlinear models' input-output relationships using GPC. The findings further suggest that geometric transformations and brightness adjustments alter the visual structure in a way that negatively affects the prediction accuracy of the classification model. This can be attributed to distributional domain shifts that differ from the models' training dataset.
Quantifying the relative influence of input transformations using Sobol indices provides valuable insights into how variations in the input affect the model's predictions. This information can assist practitioners in enhancing the robustness of models, for example, making the emblem classification model more resilient to different lighting conditions.

However, several questions remain unanswered. It remains to be validated whether the Sobol indices generalize across the entire distribution of the dataset space for a fixed input, suggesting a potential avenue for future research.
Moreover, in this study, we test only one family of polynomials, specifically the Jacobian polynomials. This leaves potential research opportunities open to exploring other polynomial families and examining their impact on the accuracy and efficiency of the surrogate model. Further investigation is required to determine the robustness of our method across different datasets and GPC configurations.

\bibliographystyle{ieeetr}
\bibliography{bib}

\end{document}